\pdfoutput=1

\documentclass[11pt]{article}

\usepackage{acl}

\usepackage{times}
\usepackage{latexsym}

\usepackage[T1]{fontenc}

\usepackage[utf8]{inputenc}

\usepackage{microtype}

\usepackage{inconsolata}

\usepackage{hyperref}       
\usepackage{url}            
\usepackage{booktabs}       
\usepackage{amsfonts}       
\usepackage{nicefrac}       
\usepackage{microtype}      
\usepackage{xcolor}         
\usepackage{comment}
\usepackage{bm}
\usepackage{enumitem}
\usepackage{amsmath}        
\usepackage{mathtools}
\usepackage{wrapfig, lipsum}
\usepackage{adjustbox}
\usepackage{dblfloatfix}    
\usepackage{tabularx}

\usepackage{graphicx}
\usepackage{multirow, booktabs}
\usepackage{subcaption}
\usepackage{makecell}
\newcommand{\tpm}{\tiny{$\pm$} }

\title{Efficient Sample-Specific Encoder Perturbations}

\author{
    Yassir Fathullah, 
    Mark J. F. Gales \\
    ALTA Institute, Department of Engineering, University of Cambridge \\
    {\tt yf286@cam.ac.uk}, 
    {\tt mjfg@eng.cam.ac.uk}
}
\begin{document}

\maketitle

\begin{abstract}
    Encoder-decoder foundation models have displayed state-of-the-art performance on a range of autoregressive sequence tasks. This paper proposes a simple and lightweight modification to such systems to control the behaviour according to a specific attribute of interest.
    %
    %
    Specifically, we show that a small proxy network can be used to find a sample-by-sample perturbation of the encoder output of a frozen foundation model to trigger the decoder to generate improved decodings. 
    This work explores a specific realization of this framework focused on improving the COMET performance of Flan-T5 on Machine Translation and the WER of Whisper foundation models on Speech Recognition. Results display consistent improvements in performance evaluated through COMET and WER respectively.
    Furthermore, experiments also show that the proxies are robust to the exact nature of the data used to train them and can extend to other domains.
\end{abstract}

\section{Introduction}


Encoder-decoder models have displayed state-of-the-art performance in a wide range of sequence tasks \cite{sutskever2014sequence} including Machine Translation (MT) \cite{transformers, mt5}, Abstractive Text Summarization \& Question Answering \cite{flan-palm} and Automatic Speech Recognition (ASR) \cite{chiu2018state}. However, the standard approach to training these systems often relies on teacher-forcing with the likelihood criteria, e.g. next token prediction of the reference sequence. While this framework has been shown successful and reliable in the tasks above, often the desired criteria are some sequence-level non-differentiable performance measures. In MT the desired criteria was the n-gram-based (Sacre)BLEU \cite{sacrebleu}, overtaken more recently by the neural-based COMET \cite{rei-comet} evaluation metric. In ASR the criteria is the word error rate (WER) measuring the rate of substitutions, insertions, and deletions between the decoded output and the reference.

Prior work has addressed the exposure bias arising from training in teacher-forcing \cite{teacher-forcing} and the loss mismatch between the likelihood and the desired sequence-level loss \cite{scheduled-sampling, professor-forcing, levenshtein-transformer, ocd, rl-nmt-study, ranzato2016seqtrain, bahdanau2017actor, wiseman-etal-2016-learning, kim-rush-2016-sequence}. 
%
%
Both of these approaches modify the training so it more closely links with how the model would be used during deployment. 
However, in the regime of large pre-trained foundation models, re-training such systems is computationally expensive and unstable. Other approaches attack this problem at inference time by merging outputs from several systems guided by appropriate sequence-level metrics \cite{sim2007consesusnetwork, kumar-byrne-2004-minimum, freitag-etal-2022-high, rosti-etal-2007-combining, rosti-etal-2007-improved, manakul-etal-2023-cued}. Whilst these approaches have shown promising gains, they rely on the use of ensembles which have significantly higher computational costs. Furthermore, they are not generalizable to any metric. 

In this paper, we propose a novel, simple and efficient approach to modify the behaviour of a single frozen pre-trained encoder-decoder foundation model. 
%
%
%
%
We show that it is possible to perturb the outputs of the encoder to trigger the decoder to produce better-performing decodings according to some pre-selected generic attribute. Unlike prior approaches, our novel proposal applies to frozen pre-trained systems and only leads to an insignificant increase in runtime. 
This paper is focused on showing the efficacy of the approach in improving the COMET performance of Flan-T5 \cite{flan-palm} on NMT and the WER performance of Whisper \cite{whisper} on ASR. Furthermore, the approach is generalizable and applicable to other attributes such as the sentiment of outputs.


\section{Background}



Reinforcement Learning (RL) and Minimum Bayes Risk (MBR) decoding have been two paradigms used to align a sequence model and minimize the impact of exposure bias. The former is traditionally used to improve the training while the latter is used to modify the decoding procedure. Both come with their own sets of advantages and disadvantages. While often leading to better performance, these approaches are often expensive to use during training and/or inference. 
%
%
%
%
%
%
In the works of \citet{bahdanau2017actor, professor-forcing}, a second network is used, either as a critic in an actor-critic framework \cite{actor-critic-1, actor-critic-2} or as a discriminator in a generative adversarial framework \cite{goodfellow2014generative}. The aim of both of these networks, although mechanically different, is to ensure the training procedure resembles the inference stage, minimising the effect of exposure bias.
On the other hand, the work of \citet{freitag-etal-2022-high} explored a post-training approach in which the Minimum Bayes Risk decoder samples many sequences from the system and chooses the sample with the lowest risk. However, the efficacy of this system is highly dependent on being able to sample a large set of outputs, a feat not possible for large pre-trained systems. Alternatively, \citet{manakul2023selfcheckgpt} applies this approach to an ensemble of similarly performing systems, without regard to the inference cost.
There is also a body of work on memory and inference-efficient adaptation of foundation models such as prefix-tuning \cite{li-liang-2021-prefix} and low-rank adaptation \cite{hu2022lora}. These approaches are effective at adapting a foundation model to a certain task at the cost of degrading other abilities. In addition, these approaches still require back-propagating through the whole foundation model making them potentially expensive to train and mainly target teacher-forcing likelihood training. Finally, the work of \citet{naps} introduced a general framework in which an encoder is extended with a small Non-Autoregressive Proxy (NAP) trained to directly capture an arbitrary metric. While only applicable to encoder-decoder systems, it showed that estimates produced by NAP systems were useful in downstream tasks. 
 

\section{Perturbations of Encoders}



The proposal's core is a flexible and efficient approach for augmenting the behaviour of an encoder on a sample-by-sample basis to trigger a better decoder performance. The starting point for such a goal will be the recently introduced Non-Autoregressive Proxy (NAP). While the original work trained the network on a specific metric and used the estimates at runtime to directly perform various downstream tasks, we will extend the view of this network to a differentiable approximation of a sequence-level metric, and use the gradients of this approximation to improve performance.

Let $\bm\phi_{\tt e}$ and $\bm\phi_{\tt d}$ represent the parameters of some encoder and decoder network, and let $\bm x$ be some input (token or embedding) sequence
%
%
At inference time we have $\bm e_{1:L} = \bm f(\bm x; \bm\phi_{\tt e})$ where $\bm e_{1:L}$ represents a sequence of $L$ encoder embeddings that are consumed by the decoder. The decoder, through an autoregressive process, produces an output sequence $\hat{\bm y} = \bm f(\bm e_{1:L}; \bm\phi_{\tt d})$. We aim to find a sample-specific perturbation $\bm \delta_{1:L}$ to the sequence of encoder outputs such that $\bar{\bm y} = \bm f(\bm e_{1:L} + \bm \delta_{1:L}; \bm\phi_{\tt d})$ gives us a higher score according to some score $\mathcal{S}$ (e.g. COMET):
\begin{align}
    \mathcal{S}(\bm y, \bar{\bm y} ; \bm x) > \mathcal{S}(\bm y, \hat{\bm y} ; \bm x)
\end{align}
where $\bm y$ is the reference sequence. To find a good perturbation we first train a lightweight Non-Autoregressive Proxy on top of the encoder to approximate the score $f(\bm e; \phi_{\tt nap}) \approx \mathcal{S}(\bm y, \hat{\bm y}; \bm x)$ where $\phi_{\tt nap}$ represent the NAP parameters. Once this is achieved, the gradient of the NAP can be used to find a good perturbation to the encoder outputs of a certain sample $\bm x$, making the approach sample specific:
\begin{align}
    \label{eq:perturb}
    \hspace{-1mm}
    \hspace{-1mm}
    \bm \delta_i = 
    \hspace{-0.3mm}
    \alpha 
    \hspace{-0.3mm}
    \left\vert \bm e_i \right\vert 
    \hspace{-0.3mm}
    \left\vert \frac{\partial f(\bm e_{1:L}; \phi_{\tt nap})}{\partial \bm e_i} \right\vert^{-1} 
    \hspace{-0.5mm}
    \frac{\partial f(\bm e_{1:L}; \phi_{\tt nap})}{\partial \bm e_i}
\end{align}
where $i = 1, \dots, L$. We normalize for the size of the gradient and the encoder L2-norms and include a hyperparameter $\alpha$ to control the perturbation size. Small perturbations will have no impact while large changes can lead to a degradation in performance. Therefore, the choice of $\alpha$ is important and is based on some validation set. Note that our approach is aimed to be a lightweight and cheap method for obtaining performance gains, and is not designed to achieve state-of-the-art performance. Furthermore, to the best of the authors' knowledge, no prior approaches exist for augmenting the behaviour of frozen encoder-decoder systems according to any criteria $\mathcal{S}$ which can be anything from COMET and SacreBLEU to the sentiment of the output.

\begin{figure*}[ht!]
    \centering
    \includegraphics[width=1.0\textwidth]{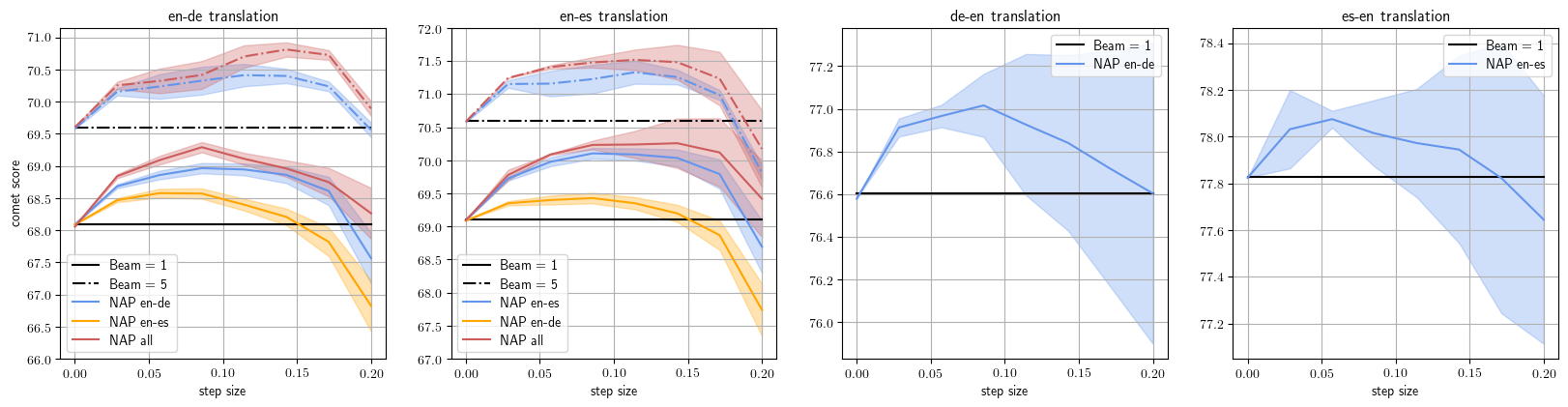}
    \caption{Modifying encoder outputs using the gradient of a NAP. This shows how the performance changes when we vary the step size $\alpha$ of the modification. The error represents 1 standard deviation. The various figures show the translation task in the title and NAPs (data each was trained on) in the legend.}
    \label{fig:improvement-nmt}
\end{figure*}
\begin{figure*}[!b]
    \centering
    \includegraphics[width=0.75\textwidth]{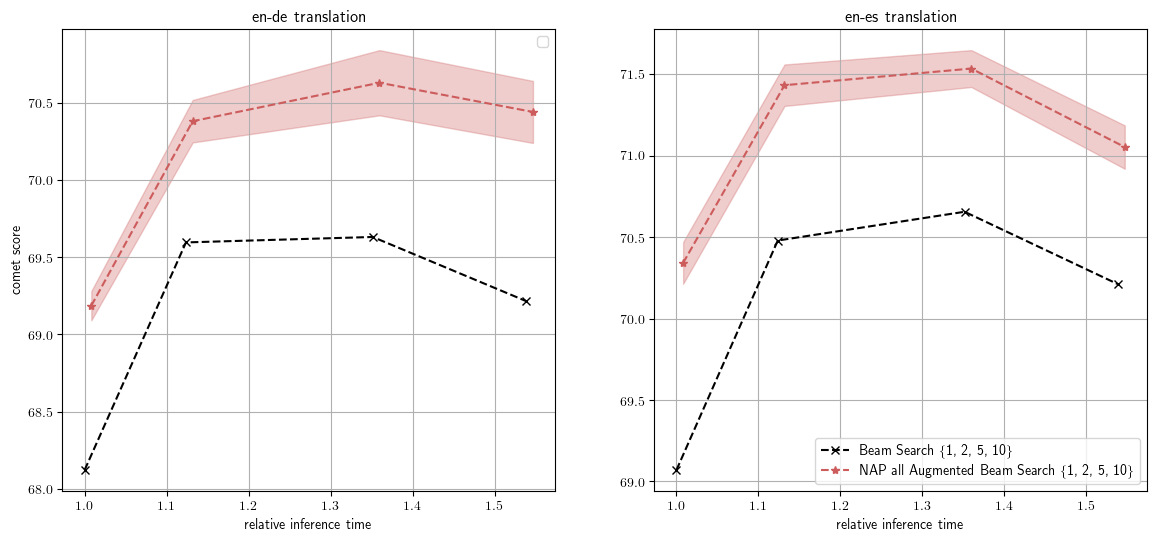}
    \caption{Modifying encoder outputs using the gradient of a NAP. This shows the overall performance of various beams using an optimal fixed $\alpha = 0.50$ based on the validation sets. The error represents 1 standard deviation.}
    \label{fig:speed}
\end{figure*}

\section{Experimental Evaluation}

The first set of experiments aim to evaluate the efficacy of the gradient perturbation approach on NMT using the pre-trained Flan-T5. We resort to the OPUS-100 \cite{zhang-etal-2020-improving} dataset and use two splits, the English to German (en$\to$de) and the English to Spanish (en$\to$es) splits, each with a training set of 1M sentence pairs. For each, we use the Flan-T5 system to decode the data using greedy search with a maximum of 128 output tokens. The decodings are scored using COMET (\texttt{Unbabel/wmt22-comet-da}) \cite{rei-comet}. Following \citet{naps} we train a small NAP on the Flan-T5 encoder to predict these COMET scores, using the Pearson correlation loss, training details provided in Appendix \ref{sapp:nmt}. Each experiment is repeated 3 times.

\begin{table}[ht!]
	\centering{}
	\begin{minipage}[t]{0.48\textwidth}%
		\begin{center}
                \caption{NAP Pearson correlation $\uparrow$.}
			\vspace{-3mm}
			\def\arraystretch{1.0}
			\small
    		\begin{tabular}{c|cc|cc}
    			\toprule
                    \multirow{2}{*}{Train} & \multicolumn{4}{c}{Eval} \\
                    & en$\to$de & en$\to$es & de$\to$en & es$\to$en \\
                    \midrule
                    en$\to$de & \textbf{75.6} {\tpm 0.1} & 54.8{\tpm 0.4} & 52.1 \tpm 1.6 & 38.7 \tpm 1.9 \\
                    en$\to$es  & 65.7 {\tpm 0.2} & \textbf{68.1} {\tpm 0.1} & 52.4 \tpm 0.4 & 35.2 \tpm 1.4\\
				\bottomrule
                \end{tabular}
			\label{tab:correlation}
		\end{center}
	\end{minipage}
\end{table}

Table \ref{tab:correlation} presents the Pearson correlation between the NAP predictions and the COMET scores on various test sets. We observe that a proxy trained on a specific translation pair can still obtain good correlation scores on other splits and reverse directions. The worst performance is shown when a proxy trained on en$\to$es pairs is evaluated on the reverse direction displaying a correlation of 35.2\%. 

\begin{figure*}[!b]
    \centering
    \includegraphics[width=1.0\textwidth]{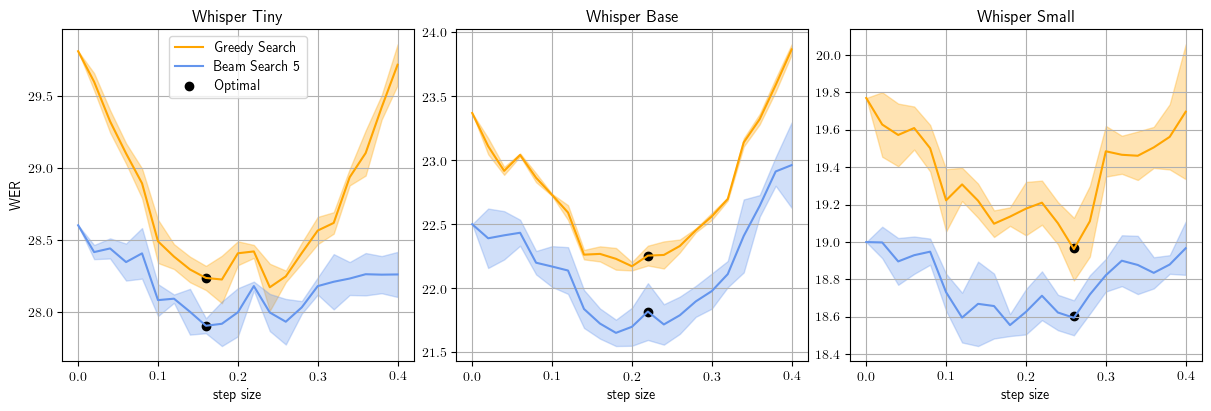}
    \caption{Modifying encoder outputs using the gradient of a NAP. This shows how the performance changes when we vary the step size $\alpha$ of the modification. The optimal point was decided based on the minimum WER found on the validation set.}
    \label{fig:improvement-asr}
\end{figure*}

Next, we take these NAPs and use the gradient with respect to the encoder outputs to perform the augmentation detailed in Equation (\ref{eq:perturb}), see Figure \ref{fig:improvement-nmt}.
This shows that the gradients derived from a lightweight NAP trained on COMET scores can be used to obtain some performance gains. While small steps $\alpha$ barely have an impact, large $\alpha$ lead to a degradation in performance showcasing the importance of finding a good $\alpha$. Furthermore, we find that although the NAPs were trained on the COMET scores of greedy decodings, they can still be used to improve beam search. We also observe that NAPs trained in a certain translation direction can still be used for other directions. For example, NAP en-es is still able to improve en-de performance.

We also investigate this for a range of Flan-T5 models, see Table \ref{tab:comet-size}. In all cases we use a fixed value of $\alpha = 0.50$. 
\begin{table}[ht!]
	\centering{}
	\begin{minipage}[t]{0.48\textwidth}%
		\begin{center}
                \caption{Flan-T5 en-de COMET performance $\uparrow$.}
			\vspace{-2mm}
			\def\arraystretch{1.0}
			\small
    		\begin{tabular}{c|ccc}
    			\toprule
                     & Small & \hspace{-2mm} Base \hspace{-2mm} & Large \\
                    \midrule
                    Greedy Search & 58.8 \tpm 0.4 & \hspace{-2mm} 64.3 \tpm 0.4 \hspace{-2mm} & 68.1 \tpm 0.3 \\
                     + NAP en-de Perturb. & 61.4 \tpm 0.5 & \hspace{-2mm} 66.0 \tpm 0.4 \hspace{-2mm} & 69.9 \tpm 0.3 \\
                    \bottomrule
                \end{tabular}
			\label{tab:comet-size}
		\end{center}
	\end{minipage}
        \vspace{-1mm}
\end{table} 
From these sets of results, it is evident that smaller models benefit more from the perturbation approach while larger more robust models show smaller gains.

Furthermore, we apply this approach to a range of different-sized beam search decodings, see Figure \ref{fig:speed}. For a small cost of performing a forward and backward pass through the small NAP network, both translation pairs show gains across all beams. Note, that the inference speed was measured using an NVIDIA A100 80GBs leading to a disproportionally cheaper runtime for larger beam sizes, since the GPU cores were not fully exhausted for a single sample. Interestingly while we observe improvements in COMET scores, the SacreBLEU score remained the same, see Table \ref{tab:sacre}. 
\begin{table}[ht!]
	\centering{}
	\begin{minipage}[t]{0.48\textwidth}%
		\begin{center}
                \caption{Flan-T5 SacreBLEU $\uparrow$.}
			\vspace{-2mm}
			\def\arraystretch{1.0}
			\small
    		\begin{tabular}{c|cc}
    			\toprule
                     & en$\to$de & en$\to$es \\
                    \midrule
                    Greedy Search & 19.8 \tpm 0.2 & 23.2 \tpm 0.3 \\
                     + NAP all Perturb. & 19.6 \tpm 0.2 & 23.3 \tpm 0.2 \\
                    \bottomrule
                \end{tabular}
			\label{tab:sacre}
		\end{center}
	\end{minipage}
\end{table} 
Upon further analysis of this phenomenon, we observed two factors contributing to this effect. The first is that the proxy augmentation would substitute certain tokens/words that have a "closer" meaning to the reference but without any n-gram overlap. The second factor is discussed in \citet{stahlberg-byrne-2019-nmt} in which translation systems often terminate prematurely. The proxy-derived augmentation of the encoder resolves this issue in a fraction of examples, but the continuation of the translation often does not overlap with the words occurring in the reference.

Finally, we repeat these experiments for the Whisper family on the AMI meeting corpus, a challenging ASR task in which systems often display very high word error rates. Following \citet{naps} we train NAPs on the number of errors in a transcription produced by Whisper \{Tiny, Base, Small\} using greedy search, see Appendix \ref{sapp:asr} for details. The performance of the augmentation is displayed in Figure \ref{fig:improvement-asr}. Four observations made in these results are: (1) the proxy gradients are beneficial for Whisper in the ASR task, (2) the validation and test sets show correlating performance, (3) the step sizes are significantly larger since the gradients from the NAP are significantly smaller and (4) the improvement is smaller for the larger Whisper systems. Concerning the last point, this behaviour was expected since larger more robust systems potentially have a smaller room for improvement. Similar experiments performed on LibriSpeech \cite{librispeech}, a relatively easier speech recognition benchmark showcased small to no gains supporting the claim that the benefits from our proposed approach is highly dependent on the task given a certain model. The details of the LibriSpeech experiments are included in Appendix \ref{app:libri}.

\section{Conclusion}

This paper presents a lightweight modification of frozen (pre-trained) encoder-decoder systems by training a small network to predict a metric and use this as a differentiable extension. The gradients of this small proxy network have been shown useful in improving the performance on a range of sequence tasks such as the Flan-T5 COMET performance on NMT and Whisper WER performance on ASR.

\section*{Limitations}


The main issue raised by \citet{naps} was that NAPs were specifically designed for encoder-decoder systems and might not be so easily extended to decoder-only systems such as standard Large Language Models \cite{touvron2023llama, touvron2023llama2, brown2020language, open2023gpt4}. Because of this limitation, our approach is also limited to encoder-decoder systems. However, many multimodal language models utilize modal-specific encoders \cite{fathullah2023prompting, fathullah2023towards, lakomkin2023end, rubenstein2023audiopalm, gong2023listen, palme}. Therefore, future work could investigate applying our approach to augmented language models by extending the modal-speific encoders with proxies to estimate the system performance at certain tasks. Furthermore, one possible limitation of our approach is its effectiveness on models that are already well performing. Experiments investigating Whisper on LibriSpeech showcase a very marginal change in WER performance which we attribute to Whisper's already good performance on the benchmark. 

\section*{Acknowledgements}

This paper reports on research supported by the Gates Cambridge Trust (grant OPP1144 from the Bill \& Melinda Gates Foundation). This research is further supported by Cambridge University Press \& Assessment (CUP\&A), a department of The Chancellor, Masters, and Scholars of the University of Cambridge.

\bibliography{anthology, custom}

\begin{thebibliography}{45}
\expandafter\ifx\csname natexlab\endcsname\relax\def\natexlab#1{#1}\fi

\bibitem[{Bahdanau et~al.(2017)Bahdanau, Brakel, Xu, Goyal, Lowe, Pineau, Courville, and Bengio}]{bahdanau2017actor}
Dzmitry Bahdanau, Philemon Brakel, Kelvin Xu, Anirudh Goyal, Ryan Lowe, Joelle Pineau, Aaron Courville, and Yoshua Bengio. 2017.
\newblock An actor-critic algorithm for sequence prediction.
\newblock In \emph{International Conference on Learning Representations (ICLR)}.

\bibitem[{Barto et~al.(1983)Barto, Sutton, and Anderson}]{actor-critic-1}
Andrew~G. Barto, Richard~S. Sutton, and Charles~W. Anderson. 1983.
\newblock Neuronlike adaptive elements that can solve difficult learning control problems.
\newblock \emph{IEEE Transactions on Systems, Man, and Cybernetics}.

\bibitem[{Bengio et~al.(2015)Bengio, Vinyals, Jaitly, and Shazeer}]{scheduled-sampling}
Samy Bengio, Oriol Vinyals, Navdeep Jaitly, and Noam Shazeer. 2015.
\newblock Scheduled sampling for sequence prediction with recurrent neural networks.
\newblock \emph{Conference on Neural Information Processing Systems}.

\bibitem[{Blondel et~al.(2020)Blondel, Teboul, Berthet, and Djolonga}]{sorting}
Mathieu Blondel, Olivier Teboul, Quentin Berthet, and Josip Djolonga. 2020.
\newblock Fast differentiable sorting and ranking.
\newblock \emph{International Conference on Machine Learning}.

\bibitem[{Brown et~al.(2020)Brown, Mann, Ryder, Subbiah, Kaplan, Dhariwal, Neelakantan, Shyam, Sastry, Askell et~al.}]{brown2020language}
Tom Brown, Benjamin Mann, Nick Ryder, Melanie Subbiah, Jared~D Kaplan, Prafulla Dhariwal, Arvind Neelakantan, Pranav Shyam, Girish Sastry, Amanda Askell, et~al. 2020.
\newblock Language models are few-shot learners.
\newblock \emph{Advances in neural information processing systems}, 33:1877--1901.

\bibitem[{Chiu et~al.(2018)Chiu, Sainath, Wu, Prabhavalkar, Nguyen, Chen, Kannan, Weiss, Rao, Gonina et~al.}]{chiu2018state}
Chung-Cheng Chiu, Tara~N Sainath, Yonghui Wu, Rohit Prabhavalkar, Patrick Nguyen, Zhifeng Chen, Anjuli Kannan, Ron~J Weiss, Kanishka Rao, Ekaterina Gonina, et~al. 2018.
\newblock State-of-the-art speech recognition with sequence-to-sequence models.
\newblock In \emph{2018 IEEE international conference on acoustics, speech and signal processing (ICASSP)}, pages 4774--4778. IEEE.

\bibitem[{Chung et~al.(2022)Chung, Hou, Longpre, Zoph, Tay, Fedus, Li, Wang, Dehghani, Brahma et~al.}]{flan-palm}
Hyung~Won Chung, Le~Hou, Shayne Longpre, Barret Zoph, Yi~Tay, William Fedus, Eric Li, Xuezhi Wang, Mostafa Dehghani, Siddhartha Brahma, et~al. 2022.
\newblock Scaling instruction-finetuned language models.
\newblock \emph{arXiv preprint arXiv:2210.11416}.

\bibitem[{Driess et~al.(2023)Driess, Xia, Sajjadi, Lynch, Chowdhery, Ichter, Wahid et~al.}]{palme}
Danny Driess, Fei Xia, Mehdi S.~M. Sajjadi, Corey Lynch, Aakanksha Chowdhery, Brian Ichter, Ayzaan Wahid, et~al. 2023.
\newblock Palm-e: An embodied multimodal language model.
\newblock In \emph{Proceedings of the 40th International Conference on Machine Learning}, ICML'23. JMLR.org.

\bibitem[{Fathullah et~al.(2023{\natexlab{a}})Fathullah, Radmard, Liusie, and Gales}]{naps}
Yassir Fathullah, Puria Radmard, Adian Liusie, and Mark J.~F. Gales. 2023{\natexlab{a}}.
\newblock Who needs decoders? efficient estimation of sequence-level attributes.
\newblock \emph{arXiv, arXiv:2305.05098}.

\bibitem[{Fathullah et~al.(2023{\natexlab{b}})Fathullah, Wu, Lakomkin, Jia, Shangguan, Li, Guo, Xiong, Mahadeokar, Kalinli et~al.}]{fathullah2023prompting}
Yassir Fathullah, Chunyang Wu, Egor Lakomkin, Junteng Jia, Yuan Shangguan, Ke~Li, Jinxi Guo, Wenhan Xiong, Jay Mahadeokar, Ozlem Kalinli, et~al. 2023{\natexlab{b}}.
\newblock Prompting large language models with speech recognition abilities.
\newblock \emph{arXiv preprint arXiv:2307.11795}.

\bibitem[{Fathullah et~al.(2023{\natexlab{c}})Fathullah, Wu, Lakomkin, Jia, Shangguan, Mahadeokar, Kalinli, Fuegen, and Seltzer}]{fathullah2023towards}
Yassir Fathullah, Chunyang Wu, Egor Lakomkin, Junteng Jia, Yuan Shangguan, Jay Mahadeokar, Ozlem Kalinli, Christian Fuegen, and Mike Seltzer. 2023{\natexlab{c}}.
\newblock Towards general-purpose speech abilities for large language models using unpaired data.
\newblock \emph{arXiv preprint arXiv:2311.06753}.

\bibitem[{Freitag et~al.(2022)Freitag, Grangier, Tan, and Liang}]{freitag-etal-2022-high}
Markus Freitag, David Grangier, Qijun Tan, and Bowen Liang. 2022.
\newblock \href {https://doi.org/10.1162/tacl_a_00491} {High quality rather than high model probability: Minimum {B}ayes risk decoding with neural metrics}.
\newblock \emph{Transactions of the Association for Computational Linguistics}, 10:811--825.

\bibitem[{Gong et~al.(2023)Gong, Luo, Liu, Karlinsky, and Glass}]{gong2023listen}
Yuan Gong, Hongyin Luo, Alexander~H Liu, Leonid Karlinsky, and James Glass. 2023.
\newblock Listen, think, and understand.
\newblock \emph{arXiv preprint arXiv:2305.10790}.

\bibitem[{Goodfellow et~al.(2014)Goodfellow, Pouget-Abadie, Mirza, Xu, Warde-Farley, Ozair, Courville, and Bengio}]{goodfellow2014generative}
Ian Goodfellow, Jean Pouget-Abadie, Mehdi Mirza, Bing Xu, David Warde-Farley, Sherjil Ozair, Aaron Courville, and Yoshua Bengio. 2014.
\newblock Generative adversarial nets.
\newblock \emph{Advances in neural information processing systems}, 27.

\bibitem[{Gu et~al.(2019)Gu, Wang, and Zhao}]{levenshtein-transformer}
Jiatao Gu, Changhan Wang, and Jake Zhao. 2019.
\newblock Levenshtein transformer.
\newblock \emph{Conference on Neural Information Processing Systems}.

\bibitem[{Hu et~al.(2022)Hu, Shen, Wallis, Allen-Zhu, Li, Wang, Wang, and Chen}]{hu2022lora}
Edward~J Hu, Yelong Shen, Phillip Wallis, Zeyuan Allen-Zhu, Yuanzhi Li, Shean Wang, Lu~Wang, and Weizhu Chen. 2022.
\newblock \href {https://openreview.net/forum?id=nZeVKeeFYf9} {Lo{RA}: Low-rank adaptation of large language models}.

\bibitem[{Kim and Rush(2016)}]{kim-rush-2016-sequence}
Yoon Kim and Alexander~M. Rush. 2016.
\newblock \href {https://doi.org/10.18653/v1/D16-1139} {Sequence-level knowledge distillation}.
\newblock In \emph{Proceedings of the 2016 Conference on Empirical Methods in Natural Language Processing}, pages 1317--1327, Austin, Texas. Association for Computational Linguistics.

\bibitem[{Kumar and Byrne(2004)}]{kumar-byrne-2004-minimum}
Shankar Kumar and William Byrne. 2004.
\newblock \href {https://aclanthology.org/N04-1022} {Minimum {B}ayes-risk decoding for statistical machine translation}.
\newblock In \emph{Proceedings of the Human Language Technology Conference of the North {A}merican Chapter of the Association for Computational Linguistics: {HLT}-{NAACL} 2004}, pages 169--176, Boston, Massachusetts, USA. Association for Computational Linguistics.

\bibitem[{Lakomkin et~al.(2023)Lakomkin, Wu, Fathullah, Kalinli, Seltzer, and Fuegen}]{lakomkin2023end}
Egor Lakomkin, Chunyang Wu, Yassir Fathullah, Ozlem Kalinli, Michael~L Seltzer, and Christian Fuegen. 2023.
\newblock End-to-end speech recognition contextualization with large language models.
\newblock \emph{arXiv preprint arXiv:2309.10917}.

\bibitem[{Lamb et~al.(2016)Lamb, Goyal, Zhang, Zhang, Courville, and Bengio}]{professor-forcing}
Alex Lamb, Anirudh Goyal, Ying Zhang, Saizheng Zhang, Aaron Courville, and Yoshua Bengio. 2016.
\newblock Professor forcing: A new algorithm for training recurrent networks.
\newblock \emph{Conference on Neural Information Processing Systems}.

\bibitem[{Li and Liang(2021)}]{li-liang-2021-prefix}
Xiang~Lisa Li and Percy Liang. 2021.
\newblock \href {https://doi.org/10.18653/v1/2021.acl-long.353} {Prefix-tuning: Optimizing continuous prompts for generation}.
\newblock In \emph{Proceedings of the 59th Annual Meeting of the Association for Computational Linguistics and the 11th International Joint Conference on Natural Language Processing (Volume 1: Long Papers)}, pages 4582--4597, Online. Association for Computational Linguistics.

\bibitem[{Manakul et~al.(2023{\natexlab{a}})Manakul, Fathullah, Liusie, Raina, Raina, and Gales}]{manakul-etal-2023-cued}
Potsawee Manakul, Yassir Fathullah, Adian Liusie, Vyas Raina, Vatsal Raina, and Mark Gales. 2023{\natexlab{a}}.
\newblock \href {https://doi.org/10.18653/v1/2023.bionlp-1.51} {{CUED} at {P}rob{S}um 2023: Hierarchical ensemble of summarization models}.
\newblock In \emph{The 22nd Workshop on Biomedical Natural Language Processing and BioNLP Shared Tasks}, pages 516--523, Toronto, Canada. Association for Computational Linguistics.

\bibitem[{Manakul et~al.(2023{\natexlab{b}})Manakul, Liusie, and Gales}]{manakul2023selfcheckgpt}
Potsawee Manakul, Adian Liusie, and Mark~JF Gales. 2023{\natexlab{b}}.
\newblock Selfcheckgpt: Zero-resource black-box hallucination detection for generative large language models.
\newblock \emph{arXiv preprint arXiv:2303.08896}.

\bibitem[{OpenAI(2023)}]{open2023gpt4}
OpenAI. 2023.
\newblock \href {https://api.semanticscholar.org/CorpusID:257532815} {Gpt-4 technical report}.
\newblock \emph{ArXiv}, abs/2303.08774.

\bibitem[{Panayotov et~al.(2015)Panayotov, Chen, Povey, and Khudanpur}]{librispeech}
Vassil Panayotov, Guoguo Chen, Daniel Povey, and Sanjeev Khudanpur. 2015.
\newblock Librispeech: An asr corpus based on public domain audio books.
\newblock \emph{International Conference on Acoustics, Speech and Signal Processing (ICASSP)}.

\bibitem[{Post(2018)}]{sacrebleu}
Matt Post. 2018.
\newblock A call for clarity in reporting {BLEU} scores.
\newblock \emph{Conference on Machine Translation: Research Papers}.

\bibitem[{Radford et~al.(2022)Radford, Kim, Xu, Brockman, McLeavey, and Sutskever}]{whisper}
Alec Radford, Jong~Wook Kim, Tao Xu, Greg Brockman, Christine McLeavey, and Ilya Sutskever. 2022.
\newblock Robust speech recognition via large-scale weak supervision.
\newblock \emph{arXiv, arXiv:2212.04356}.

\bibitem[{Ranzato et~al.(2016)Ranzato, Chopra, Auli, and Zaremba}]{ranzato2016seqtrain}
Marc'Aurelio Ranzato, Sumit Chopra, Michael Auli, and Wojciech Zaremba. 2016.
\newblock Sequence level training with recurrent neural networks.
\newblock In \emph{International Conference on Learning Representations (ICLR)}.

\bibitem[{Rei et~al.(2020)Rei, Stewart, Farinha, and Lavie}]{rei-comet}
Ricardo Rei, Craig Stewart, Ana~C Farinha, and Alon Lavie. 2020.
\newblock Comet: A neural framework for mt evaluation.
\newblock \emph{Association for Computational Linguistics}.

\bibitem[{Rosti et~al.(2007{\natexlab{a}})Rosti, Ayan, Xiang, Matsoukas, Schwartz, and Dorr}]{rosti-etal-2007-combining}
Antti-Veikko Rosti, Necip~Fazil Ayan, Bing Xiang, Spyros Matsoukas, Richard Schwartz, and Bonnie Dorr. 2007{\natexlab{a}}.
\newblock \href {https://aclanthology.org/N07-1029} {Combining outputs from multiple machine translation systems}.
\newblock In \emph{Human Language Technologies 2007: The Conference of the North {A}merican Chapter of the Association for Computational Linguistics; Proceedings of the Main Conference}, pages 228--235, Rochester, New York. Association for Computational Linguistics.

\bibitem[{Rosti et~al.(2007{\natexlab{b}})Rosti, Matsoukas, and Schwartz}]{rosti-etal-2007-improved}
Antti-Veikko Rosti, Spyros Matsoukas, and Richard Schwartz. 2007{\natexlab{b}}.
\newblock \href {https://aclanthology.org/P07-1040} {Improved word-level system combination for machine translation}.
\newblock In \emph{Proceedings of the 45th Annual Meeting of the Association of Computational Linguistics}, pages 312--319, Prague, Czech Republic. Association for Computational Linguistics.

\bibitem[{Rubenstein et~al.(2023)Rubenstein, Asawaroengchai, Nguyen, Bapna, Borsos, Quitry, Chen, Badawy, Han, Kharitonov et~al.}]{rubenstein2023audiopalm}
Paul~K Rubenstein, Chulayuth Asawaroengchai, Duc~Dung Nguyen, Ankur Bapna, Zal{\'a}n Borsos, F{\'e}lix de~Chaumont Quitry, Peter Chen, Dalia~El Badawy, Wei Han, Eugene Kharitonov, et~al. 2023.
\newblock Audiopalm: A large language model that can speak and listen.
\newblock \emph{arXiv preprint arXiv:2306.12925}.

\bibitem[{Sabour et~al.(2019)Sabour, Chan, and Norouzi}]{ocd}
Sara Sabour, William Chan, and Mohammad Norouzi. 2019.
\newblock Optimal completion distillation for sequence learning.
\newblock \emph{International Conference on Learning Representations (ICLR)}.

\bibitem[{Sim et~al.(2007)Sim, Byrne, Gales, Sahbi, and Woodland}]{sim2007consesusnetwork}
K.~C. Sim, W.~J. Byrne, M.~J.~F. Gales, H.~Sahbi, and P.~C. Woodland. 2007.
\newblock Consensus network decoding for statistical machine translation system combination.
\newblock \emph{2007 IEEE International Conference on Acoustics, Speech and Signal Processing - ICASSP '07}.

\bibitem[{Stahlberg and Byrne(2019)}]{stahlberg-byrne-2019-nmt}
Felix Stahlberg and Bill Byrne. 2019.
\newblock \href {https://doi.org/10.18653/v1/D19-1331} {On {NMT} search errors and model errors: Cat got your tongue?}
\newblock In \emph{Proceedings of the 2019 Conference on Empirical Methods in Natural Language Processing and the 9th International Joint Conference on Natural Language Processing (EMNLP-IJCNLP)}, pages 3356--3362, Hong Kong, China. Association for Computational Linguistics.

\bibitem[{Sutskever et~al.(2014)Sutskever, Vinyals, and Le}]{sutskever2014sequence}
Ilya Sutskever, Oriol Vinyals, and Quoc~V Le. 2014.
\newblock Sequence to sequence learning with neural networks.
\newblock \emph{Advances in neural information processing systems}, 27.

\bibitem[{Sutton(1984)}]{actor-critic-2}
Richard~Stuart Sutton. 1984.
\newblock Temporal credit assignment in reinforcement learning.

\bibitem[{Touvron et~al.(2023{\natexlab{a}})Touvron, Lavril, Izacard, Martinet, Lachaux, Lacroix, Rozi{\`e}re, Goyal, Hambro, Azhar et~al.}]{touvron2023llama}
Hugo Touvron, Thibaut Lavril, Gautier Izacard, Xavier Martinet, Marie-Anne Lachaux, Timoth{\'e}e Lacroix, Baptiste Rozi{\`e}re, Naman Goyal, Eric Hambro, Faisal Azhar, et~al. 2023{\natexlab{a}}.
\newblock Llama: Open and efficient foundation language models.
\newblock \emph{arXiv preprint arXiv:2302.13971}.

\bibitem[{Touvron et~al.(2023{\natexlab{b}})Touvron, Martin, Stone, Albert, Almahairi, Babaei, Bashlykov, Batra, Bhargava, Bhosale et~al.}]{touvron2023llama2}
Hugo Touvron, Louis Martin, Kevin Stone, Peter Albert, Amjad Almahairi, Yasmine Babaei, Nikolay Bashlykov, Soumya Batra, Prajjwal Bhargava, Shruti Bhosale, et~al. 2023{\natexlab{b}}.
\newblock Llama 2: Open foundation and fine-tuned chat models.
\newblock \emph{arXiv preprint arXiv:2307.09288}.

\bibitem[{Vaswani et~al.(2017)Vaswani, Shazeer, Parmar, Uszkoreit, Jones, Gomez, Kaiser, and Polosukhin}]{transformers}
Ashish Vaswani, Noam Shazeer, Niki Parmar, Jakob Uszkoreit, Llion Jones, Aidan~N Gomez, {\L}ukasz Kaiser, and Illia Polosukhin. 2017.
\newblock Attention is all you need.
\newblock \emph{Advances in neural information processing systems}, 30.

\bibitem[{Williams and Zipser(1989)}]{teacher-forcing}
Ronald~J. Williams and David Zipser. 1989.
\newblock A learning algorithm for continually running fully recurrent neural networks.
\newblock \emph{Neural Computation}.

\bibitem[{Wiseman et~al.(2016)Wiseman, Rush, and Shieber}]{wiseman-etal-2016-learning}
Sam Wiseman, Alexander~M. Rush, and Stuart~M. Shieber. 2016.
\newblock \href {https://doi.org/10.18653/v1/N16-1114} {Learning global features for coreference resolution}.
\newblock In \emph{Proceedings of the 2016 Conference of the North {A}merican Chapter of the Association for Computational Linguistics: Human Language Technologies}, pages 994--1004, San Diego, California. Association for Computational Linguistics.

\bibitem[{Wu et~al.(2018)Wu, Tian, Qin, Lai, and Liu}]{rl-nmt-study}
Lijun Wu, Fei Tian, Tao Qin, Jianhuang Lai, and Tie-Yan Liu. 2018.
\newblock A study of reinforcement learning for neural machine translation.
\newblock \emph{Conference on Empirical Methods in Natural Language Processing}.

\bibitem[{Xue et~al.(2021)Xue, Constant, Roberts, Kale, Al-Rfou, Siddhant, Barua, and Raffel}]{mt5}
Linting Xue, Noah Constant, Adam Roberts, Mihir Kale, Rami Al-Rfou, Aditya Siddhant, Aditya Barua, and Colin Raffel. 2021.
\newblock mt5: A massively multilingual pre-trained text-to-text transformer.
\newblock In \emph{Proceedings of the 2021 Conference of the North American Chapter of the Association for Computational Linguistics: Human Language Technologies}, pages 483--498.

\bibitem[{Zhang et~al.(2020)Zhang, Williams, Titov, and Sennrich}]{zhang-etal-2020-improving}
Biao Zhang, Philip Williams, Ivan Titov, and Rico Sennrich. 2020.
\newblock \href {https://doi.org/10.18653/v1/2020.acl-main.148} {Improving massively multilingual neural machine translation and zero-shot translation}.
\newblock In \emph{Proceedings of the 58th Annual Meeting of the Association for Computational Linguistics}, pages 1628--1639, Online. Association for Computational Linguistics.

\end{thebibliography}
\newpage

\begin{figure*}[h!]
	\centering
	\includegraphics[width=0.60\textwidth]{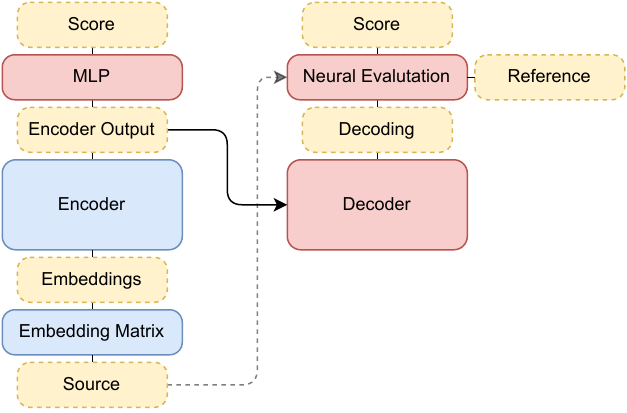}
	\caption{The proxy setup: Preemptively obtain an approximate score for the system performance through the encoder, circumventing the autoregressive decoding. The score can be differentiated with respect to encoder outputs in order to augment the system behaviour.}
	\label{fig:nap}
\end{figure*}

\vspace*{25cm}
\appendix

\section{Setup \& Training Details}
\label{app:setup}

This section will cover the details of the experiments performed in this paper. See Figure \ref{fig:nap} for a visual setup of the approach. In all experiments the MLP/proxy network consist of an attention layer with a single trainable query in order to pool the encoder output sequence, followed by three linear layers, $\mathbb{R}^{d_{\tt model}} \to \mathbb{R}^{d_{\tt ffn}}$, $\mathbb{R}^{d_{\tt ffn}} \to \mathbb{R}^{d_{\tt ffn}}$ and $\mathbb{R}^{d_{\tt ffn}} \to \mathbb{R}$, where $d_{\tt model}$ is the model dimension following the convention in \citet{transformers} and $d_{\tt ffn}$ is the feed forward dimension. The parameter sizes of the NAPs are the same as in \citet{naps}, see Table \ref{tab:sizes}. Note that although the proxy network on top of the encoder can have a relatively large size it is significantly faster since it pools the sequence of vectors and the 3 linear layers only operate on a single vector.

\begin{table}[ht!]
	\centering{}
	\begin{minipage}[t]{0.48\textwidth}%
		\begin{center}
			\caption{NAP parameter sizes}
			\vspace{-2mm}
			\def\arraystretch{1.0}
			\begin{tabular}{c|cc}
				\toprule
				Name & Model & Proxy \\
				\midrule
				Flan-T5 Large & 737.7M & 20.9M \\
				\midrule
				Whisper Tiny & 37.8M & 3.6M \\
				Whisper Base  & 72.6M & 6.3M \\
				Whisper Small & 241.7M & 14.2M \\
				\bottomrule
			\end{tabular}
			\label{tab:sizes}
		\end{center}
	\end{minipage}
\end{table} 

\subsection{Neural Machine Translation}
\label{sapp:nmt}

We follow the training details provided by \citet{naps} as closely as possible. We generated COMET scores  from  Flan-T5 Large on the training set of OPUS-100 and used them to train NAP models. We used the Pearson Correlation loss since \citet{naps} found a small difference in performance between this and the smooth extension to the Spearman Rank loss \cite{sorting}. All experiments used a learning rate of 0.0001, with a maximum batch size and training was stopped when performance did not improve after an epoch.

\subsection{Automatic Speech Recognition}
\label{sapp:asr}

Similar to the section above: We generated WER scores from Whisper \{Tiny, Base, Small\} on the training set of AMI-IHM and used them to train NAP models. We used the Pearson Correlation loss. All experiments used a learning rate of 0.0001, with a maximum batch size and training was stopped when performance did not improve after an epoch.

\begin{figure*}[!b]
    \centering
    \includegraphics[width=1.0\textwidth]{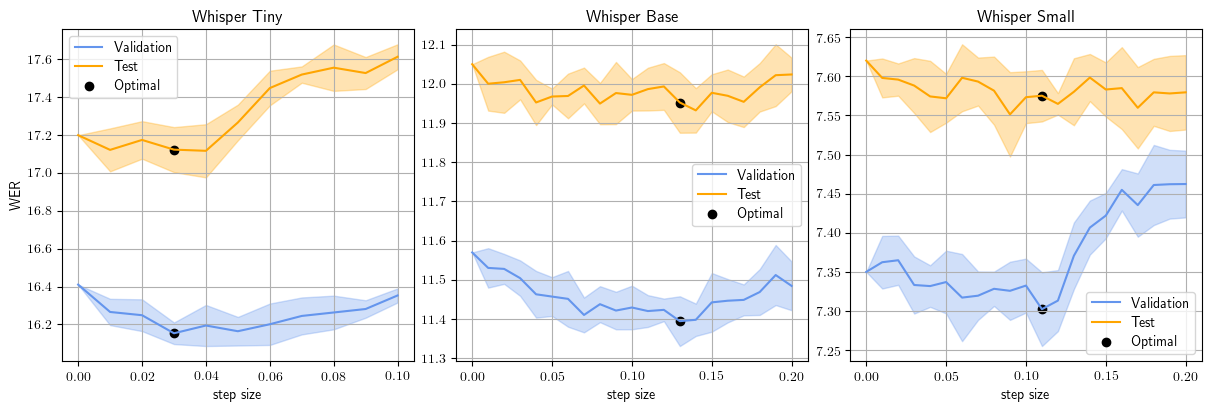}
    \caption{LibriSpeech other: Modifying encoder outputs using the gradient of a NAP. This shows how the performance changes when we vary the step size $\alpha$ of the modification. The optimal point was decided based on the minimum WER found on the validation set.}
    \label{fig:libri-improvement-asr}
\end{figure*}

\section{Whisper on LibriSpeech}
\label{app:libri}

Our experiments on the AMI corpus showed that perturbing the encoder outputs on a sample-by-sample basis could lead to WER performance improvements. However, the results in Figure \ref{fig:libri-improvement-asr} paint a different story for LibriSpeech other sets. All systems showcase a much smaller improvement and the gain is smaller for the larger Whisper systems. Since LibriSpeech is comparatively an easier benchmark, a corpus of clearly read audio books, we expect that the lack of performance improvements is due to the simplicity of the task. The Whisper systems are already performing well. AMI on the other hand represent a meeting corpus which does involve speakers speaking simultaneously and more challenging scenarios.

\end{document}